\documentclass{article}

\usepackage[final]{neurips_2021}

\usepackage[utf8]{inputenc} 
\usepackage[T1]{fontenc}    
\usepackage{hyperref}       
\usepackage{url}            
\usepackage{booktabs}       
\usepackage{amsfonts}       
\usepackage{nicefrac}       
\usepackage{microtype}      
\usepackage{xcolor}         

\usepackage{graphicx}

\title{Leveraging machine learning for less developed languages: Progress on Urdu text detection}

\author{%
  Hazrat Ali\thanks{The work was performed at COMSATS University Islamabad, Abbottabad Campus, Abbottabad, Pakistan. The dataset can be requested by email.} \\
  College of Science and Engineering, Hamad Bin Khalifa University, Doha, Qatar. \\
  \texttt{hazrat.ali@live.com} \\
}

\begin{document}

\maketitle

\begin{abstract}
Text detection in natural scene images has applications for autonomous driving, navigation help for elderly and blind people. However, the research on Urdu text detection is usually hindered by lack of data resources. We have developed a dataset of scene images with Urdu text. We present the use of machine learning methods to perform detection of Urdu text from the scene images. We extract text regions using channel enhanced Maximally Stable Extremal Region (MSER) method. First, we classify text and noise based on their geometric properties. Next, we use a support vector machine for early discarding of non-text regions. To further remove the non-text regions, we use  histogram of oriented gradients (HoG) features obtained and train a second SVM classifier. This improves the overall performance on text region detection within the scene images. To support research on Urdu text, We aim to make the data freely available for research use. We also aim to highlight the challenges and the research gap for Urdu text detection.
\end{abstract}

\section{Introduction}
Text detection in natural scene images aids in robot navigation, autonomous driving and building navigation tools for elderly people \citep{yuan2016incremental, iqbal2014bayesian}. However, research on text detection in natural scene images has made progress for developed languages only. For less developed languages, the progress is usually hindered by lack of data that is needed for training of machine learning models \citep{shahab2011icdar}. 
 Urdu is the national language of Pakistan. It is one of the major languages of the world spoken in 20 nations of the world. It is written from right to left. Some of the languages which partially share writing styles with Urdu are Arabic and Persian. However, Urdu has more number of characters and it contains additional diacritics than Arabic \citep{khan2012efficient, ali2020pioneer, ali2016urdu}. 

\textbf{Related Work:} Jamil et al., \citep{jamil2011edge} proposed edge-based features for Urdu text localization in captioned text video images. The proposed approach for Urdu text detection in captioned text videos is based on edge-based segmentation and few heuristic rules. The detected text regions are then segmented from the background. A dataset of Urdu captioned text video images is proposed by Raza et al., containing 1000 images of videos \citep{raza2012database}. There has been an attempt reported for Urdu handwritten characters recognition by \citep{ali2020pioneer}. However, this work addresses individual characters recognition on clean background and text detection in natural scene images is not addressed. Recently, dataset for Urdu text in scene images has been proposed for character extractions and recognition \citep{chandio2018character}. The dataset has 600 Urdu text scene images. Text region is manually cropped from these images and made into a dataset of 18000 Urdu characters of 48×48 dimensions for training and testing purpose. Different classifiers are trained with HOG features and efficiency of these classifiers on the proposed dataset is evaluated. Another relevant work on Urdu text recognition is reported by \citep{arafat2020IEEE}. The authors have used a deep residual network (ResNet18) for recognition of Urdu text. However, their main contribution is for recognition of text in outdoor images with synthetic text. The synthetic text is a computer-generated text with uniform fonts and size. The results for text recognition in natural scene images are limited to 76.6\% accuracy rates on a test set of 110 images.

Literature review reveals to us that progress on Urdu text detection in natural scene images is very limited. Similarly, there seems to be no publicly available data for Urdu text in natural scene images. So, one key contribution of our work is a dataset that is available publicly for research use. In addition, the proposed method provides baseline results which will motivate the development of more advanced methods for Urdu text detection. 

In the proposed method, we use a two-stage filtering approach to discard the non-text candidate regions. Firstly, we use geometric properties to classify text and non-text regions. Secondly, we use an SVM classifier to identify non-text regions which could not be removed by the geometric filtering. 
Then, we link the text regions using the vertical and horizontal distances between the centroids of the detected regions. Finally, we extract HOG features from the linked regions and use another SVM classifier to remove any left-over non-text regions. 

Our contributions in this work are: 
\begin{itemize}
    \item We report a new dataset for Urdu text images in natural scene images. The dataset has 500 unique images with Urdu text in diverse style and size. 
    \item We report preliminary results for Urdu text detection.These results will provide a good baseline to advance the research on Urdu text detection. 
    
\end{itemize}

\section{Dataset}
\label{sec:dataset}
\begin{figure}[ht!]
\centering
\includegraphics[width=0.7\linewidth]{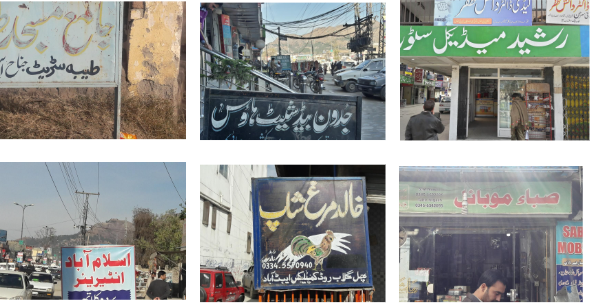}
\caption{Sample images from the dataset. Examples are chosen randomly.}
\label{fig:sampleimages}
\end{figure}

We present a dataset of 500 scene images with Urdu text. The images are recorded with a resolution of 1080 $\times$ 800. The images include recordings at different times of the day under varying light conditions. Figure \ref{fig:sampleimages} shows few of the sample images from the dataset. In our work, we use 100 images for test set and assign the rest of the images to the training set. Since we need both text and non-text regions to train the classifier, we manually crop patches from the images to get 7000+ text regions and 14000+ non-text regions. The pathces are of size 42 $\times$ 46. 

\begin{figure}[ht!]
\centering
\includegraphics[width=0.9\linewidth]{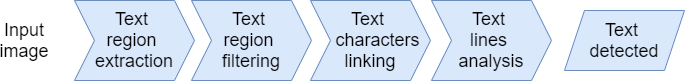}
\caption{The overall flow for the text detection. The four key modules are text region extraction, text region filtering, text character linking, and text lines analysis.}
\label{fig:workflow}
\end{figure}

\section{Methodology}
The overall workflow for the proposed model has four main working modules, as shown in Figure \ref{fig:workflow}. These four modules are; text region extraction, text region filtering, text character linking, text lines analysis. We discuss each module below. 

\textbf{Text Region Extraction:} We use the channel enhanced MSER method for extraction of text regions. 
MSER extracts the connected parts of an image with uniform intensity. The key features extracted by MSER are: stroke width, eccentricity, Euler number, and aspect ratio. Since the performance of MSER for color images (RGB images) may suffer, a channel enhanced MSER method was proposed by Yan et al. \citep{yan2017effective}. In our work, we use the channel enhanced MSER method for region proposals on text in the image. The MSER method is applied to the three channels in the image and the detection results from the three channels are combined to get the overall region proposals. 

\textbf{Text Region Filtering:} Following the text region extraction, there are a lot of false positives. We then remove the false positive using geometric filtering and SVM filtering. (a) The \emph{geometric filtering} uses geometric properties to discard non-text regions. The geometric filtering applies heuristic rules to discard the non-text regions.  These include stroke width variation, aspect ratio, eccentricity, solidity, extent, and Euler number \citep{brooks2017exploring}. (b) The \emph{SVM Filtering} is then used to filter out the remaining false positives. We train an SVM classifier with 23000 training images of size 42X46 obained from the patches. These images have 7000+ text regions and 14000+ non-text regions. The train SVM classifier is then used discard the non-text regions that were missed by the geometric filtering stage. At this stage, most of the non-text regions are discarded but few false-positive candidates are left.

\begin{figure}[!th]
\centering
\includegraphics[width=0.8\linewidth]{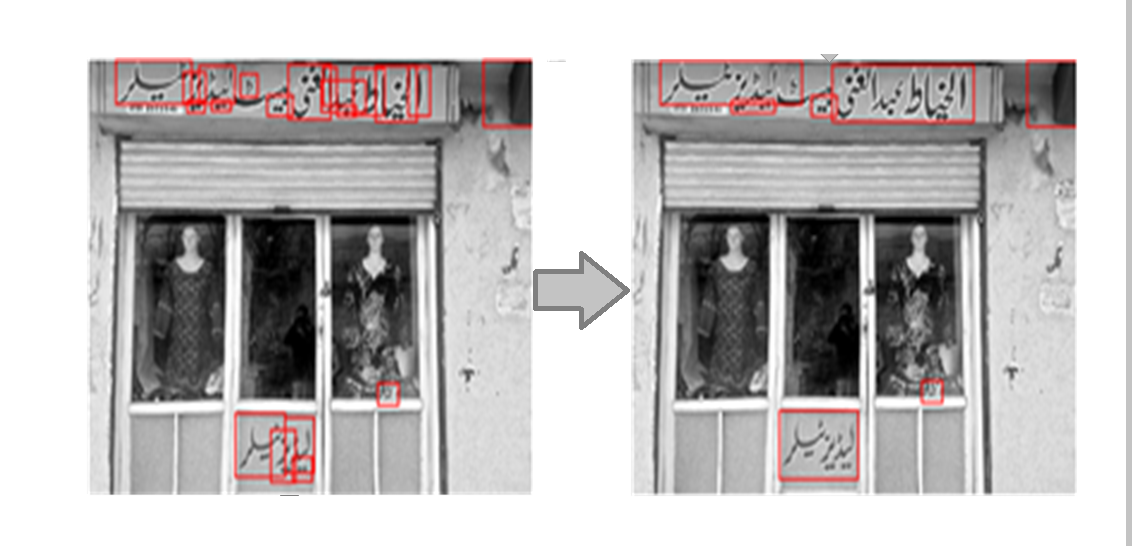}
\caption{The image on the left shows the outcome after the SVM filtering. The image on the right shows the output for the text character linking once the regions are merged into text lines.}
\label{fig:regionmerging}
\end{figure}

\textbf{Text Characters Linking:} In this module, the extracted text characters are put into pairs and lines are linked based on the centroids of each region. The centroids are joints if their vertical difference is not more than the height of the smaller character and their horizontal difference  is not more than twice the height of the larger region. 
In this way, only those text regions are joint which are closely located. Those regions which fulfil the criteria for characters linking are then merged into text lines, as shown in Figure \ref{fig:regionmerging}.


\textbf{Text Lines Analysis}: The geometric filtering followed by SVM filtering discard majority of the non-text regions. Further refinement is achieved through the text character linking module. The text character linking gives us a high recall measure (less number of false negatives) but a low precision (false positives exist). This implies that non-text regions are also detected as text regions. To overcome this, the detected text lines are further classified by using another SVM classifier trained with histogram of oriented gradients (HOG) features.

\begin{figure}[!ht]
\centering
\includegraphics[width=0.6\linewidth]{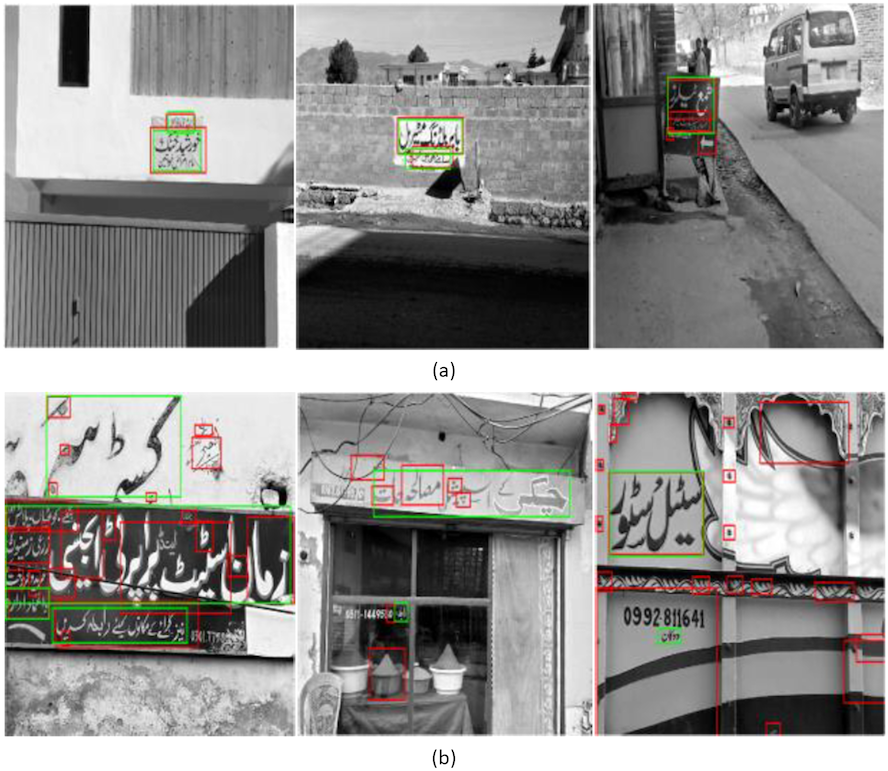}
\caption{(a) Example images for good text detection on test set images. (b) Example images for bad text detection on test set images.}
\label{fig:goodbadimages}
\end{figure}
\section{Results}
\label{sec:results}
We use a test set of 100 images to evaluate the performance. The evaluation is done by calculating the overlap ratio between the detected text region (DT) and ground truth region (GT) (also see \citep{yan2017effective}). The overlap ratio is defined as: 

\begin{equation}
\centering
    Ratio_{overlap} = \frac{Area(GT \cap DT)}{Area(GT \cup DT)}
\end{equation}

A detection is considered as a true positive if the $Ratio_{overlap} > 0.5$ and vice-versa.  

We train several SVM model using RBF kernel, and polynomial kernels of different sizes. We choose the polynomial kernel as it gave us the best detection performance. For the polynomial kernel SVM, we use polynomial of degree 3, 4, 5 and 6. We notice a performance degradation for polynomials beyond degree 5 due to over-fitting of the classifier model. The precision, recall and F-measure are reported for the different polynomial kernels in Table \ref{tab:tab4}.
Figure \ref{fig:goodbadimages} shows some samples of good and bad detection results.

\subsection{Limitations} 
\label{sec:limitations}
This work has several limitations. The dataset is small and more images need to be added. Results are preliminary and more robust machine learning models can be trained and evaluated. More challenging scenarios for scene images such as vertical orientation of text have not been studied in this work. 

\subsection{Potential Negative Impacts}
\label{sec:impacts}
Consequences of failure of the system: Overlying on the automatic detection algorithm may result in miss-reading the labels and text. False positive will put the user at risk.  \\ 
Bias in data: The data collected in this work may not be fully representative as the Urdu text in natural scene images is highly diverse with non-uniform styles and fonts.

\begin{table}[!ht]
\centering
\caption{\label{tab:tab4}Performnace of SVM with different polynomial kernels}
\begin{tabular}{lllll}
Order of Polynomial	& Precision (p)	& Recall (r)	& F-measure	& Accuracy\\ \hline
Degree 3 &	0.8928	& 0.8831	& 0.8879	& 0.8880\\ 
Degree 4 &	0.9011	& 0.8850	& 0.8930	& 0.8935\\
Degree 5 & 	0.9077	& 0.8870	& 0.8972	& 0.8980\\
Degree 6 & 	0.9079	& 0.8791	& 0.8933	& 0.8945\\
\hline
\end{tabular}
\end{table}

\section{Conclusion}
\label{sec:conclusion}
This paper presented a dataset of 500 images for Urdu text in natural scene images. The dataset can be obtained freely for research use. Furthermore, this paper reported initial results for text detection using MSER features and SVM classifier. The results reported in terms of F score show the potential of machine learning methods to automatically extract and detect Urdu text in natural scene images. This work has applications in navigation aid devices for individuals with special needs, elderly individuals and autonomous driving on roads with Urdu navigation signboards.
We believe that this work will be useful to further advance the research on Urdu text detection in scene images. Research community is welcomed to build on top of this dataset by (i) adding more images to the dataset with challenging orientations, (ii) developing more advanced machine learning and deep learning methods for text detection. 




\bibliographystyle{plainnat} %
\bibliography{ref}

\end{document}